\documentclass[journal]{IEEEtran} 

\usepackage{cite}
\usepackage{booktabs}
\usepackage[pdftex]{graphicx}
\usepackage{amsmath}
\usepackage{url}
\usepackage{hyperref}
\usepackage{color}
\usepackage{multirow}

\usepackage[caption=false,font=footnotesize]{subfig}

\begin{document}
	%
	\title{Exploiting ConvNet Diversity\\for Flooding Identification}
	%
	%
	%
	
	\author{Keiller~Nogueira,
		Samuel~G.~Fadel,
		\'{I}caro~C.~Dourado,
		Rafael~de~O.~Werneck,
		Javier~A.~V.~Mu\~{n}oz,
		Ot\'{a}vio~A.~B.~Penatti,
		Rodrigo~T.~Calumby,
		Lin~Tzy~Li,
		Jefersson~A.~dos~Santos,
		and Ricardo~da~S.~Torres
		\thanks{K.~Nogueira, and J.~A.~dos~Santos are with the Department of Computer Science, Federal University of Minas Gerais, Brazil email: \{keiller.nogueira, jefersson\}@dcc.ufmg.br}%
		\thanks{S.~G. Fadel, \'{I}.~C. Dourado, R.~O.~Werneck, J.~A.~V.~Mu\~{n}oz, L.~T.~Li, and R.~da~S.~Torres are with the Institute of Computing, University of Campinas, Brazil}%
		\thanks{L.~T.~Li, and O. A. B. Penatti are from SAMSUNG Research Institute Brazil, Campinas, Brazil}%
		\thanks{R.~T.~Calumby is with the Technology Department, University of Feira de Santana, Brazil}
	}
	
	\markboth{Nogueira \MakeLowercase{\textit{et al.}}}%
	{Nogueira \MakeLowercase{\textit{et al.}}}
	
	\maketitle
	
	\begin{abstract}
		Flooding is the world's most costly type of natural disaster in terms of both economic losses and human causalities.
		A first and essential procedure towards flood monitoring is based on identifying the area most vulnerable to flooding, which gives authorities relevant regions to focus.
		In this work, we propose several methods to perform flooding identification in high-resolution remote sensing images using deep learning.
		Specifically, some proposed techniques are based upon unique networks, such as dilated and deconvolutional ones, while other was conceived to exploit diversity of distinct networks in order to extract the maximum performance of each classifier.
		Evaluation of the proposed methods were conducted in a high-resolution remote sensing dataset. 
		Results show that the proposed algorithms outperformed state-of-the-art baselines, providing improvements ranging from 1 to 4\% in terms of the Jaccard Index.
	\end{abstract}
	

	\IEEEpeerreviewmaketitle
	
	\section{Introduction} \label{sec:intro}

	Natural disaster monitoring is a fundamental task to create prevention strategies, as well as to help authorities act in the control of damages, coordinate rescues, and assist victims.
	Among natural hazards, flooding is possibly the most extensive and devastating one, destroying buildings, and threatening human lives~\cite{NOAAFlooding}.
	All these consequences make such events to be considered as the world's most costly type of natural disaster in terms of both economic losses and human causalities~\cite{CRED}.

	Although extremely important, floods are difficult to monitor, because they are highly dependent on several local conditions, such as precipitation, slope of terrain, drainage network, protective structures, and land cover~\cite{klemas2014remote}.
	A first and essential step towards such monitoring is based on identifying areas most vulnerable to flooding, helping authorities to focus on such regions while monitoring inundations.

	Remotely sensed data may play a crucial role in the identification of such areas, since it allows the capture of whole inundated regions, allowing a better understanding of what and how they are being flooded.
	Because of the importance of such task, a subtask (called Flood-Detection in Satellite Images) of the 2017 Multimedia Satellite Task~\cite{mediaeval2017satelliteOverview}, which was part of the traditional MediaEval Benchmark, was proposed to leverage the development of methods for identifying flooding areas in high-resolution remote sensing images.
	In this paper, we present our proposed methods, which won the aforementioned task, to automatically identify flooding areas in high-resolution remote sensing images using deep learning paradigm.

	Although there are a number of previous works~\cite{pulvirenti2014flood,d2016bayesian} that performed flooding detection using remote sensing data in conjunction with elevation maps, augmenting the amount and variety of information available for an effective prediction, our work is one of the first specifically focused on identifying flooding areas on only high resolution imagery using deep learning-based approaches. 
	Furthermore, some works~\cite{pekel2016high,isikdogan2017surface} performing surface water segmentation may be suitable for flooding detection, given the high similarity between these tasks.
	However, the focus of this paper is exclusively on \textbf{flooding area identification} from satellite images took during and shortly after a flood event.

	As introduced, the proposed methods to perform flooding identification employ a resurgent area called deep learning~\cite{goodfellow2016DLbook}.
	Methods from this area, commonly represented as multi-layered neural networks, are able to learn both the features and the classifier in a unified manner, adjusting themselves to better represent the characteristics of the data and their labels.
	A specific deep learning method, called Convolutional (Neural) Networks (ConvNets)~\cite{goodfellow2016DLbook}, is the most popular for learning visual features in computer vision applications, including remote sensing~\cite{nogueira2015improving,nogueira2017towards}. 
	This type of network relies on the natural stationary property of an image, i.e., the statistics of one part of the image are assumed to be the same as those of any other part.
	Furthermore, ConvNets can be considered as an inherently multiscale approach since they usually obtain different levels of abstraction for the data, ranging from local low-level information in the initial layers (e.g., corners and edges), to more semantic descriptors, mid-level information (e.g., object parts) in intermediate layers, and high-level information (e.g., whole objects) in the final layers.
	
	Supported by these advantages, we introduce several approaches to identify flooding areas of remote sensing images exploiting ConvNets.
	Firstly, four networks have been proposed based on distinct properties, including:
	(i) dilated convolutions~\cite{YuKoltun2016}, which, unlike standard ConvNets, process the input without downsampling it, and
	(ii) deconvolution layers, employed similar to the SegNet~\cite{badrinarayanan2015segnet}, in which a coarse feature map is upsampled outputting a dense map with the same resolution of the original image.
	Then, each network is analyzed individually and in combination (with an ablation study), allowing a better understanding of their diversity, which resulted in the proposal of a combination method that aims to exploit the complementarity of these networks.
	In summary, the main contributions of the paper are the introduction of distinct network architectures, the analysis of their diversity and the proposal of combination methods that try to exploit the complementary views of different networks.
	Results of the proposed methods represent the state of the art, in terms of the Jaccard Index, in a remote-sensing-based flooding detection task.
	These results made us the winner of the Flood-Detection in Satellite Images, a subtask of 2017 Multimedia Satellite Task~\cite{mediaeval2017satelliteOverview}.

	\section{Proposed Methods} \label{sec:proposed_method_prop}

	This section introduces the network architectures (Section~\ref{sec:archs}), as well as the combination method (Section~\ref{sec:comb}) proposed to perform flooding identification. 
	
	\subsection{Network Architectures} \label{sec:archs}
	
\begin{figure}[!t]
		\centering
		\subfloat[Dilated ConvNet \#1]{
			\includegraphics[width=\columnwidth]{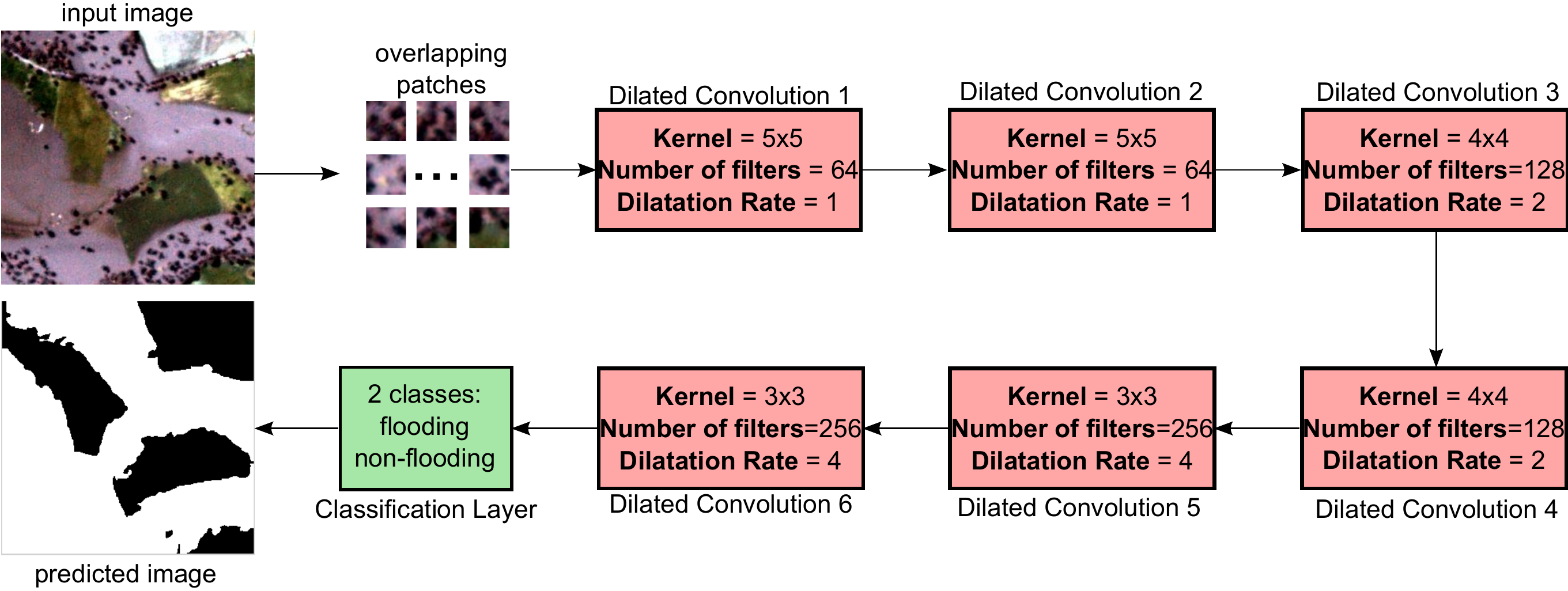}
			\label{dilated1}
		}
		\hspace{1mm}
		\subfloat[Dilated ConvNet \#2]{
			\includegraphics[width=\columnwidth]{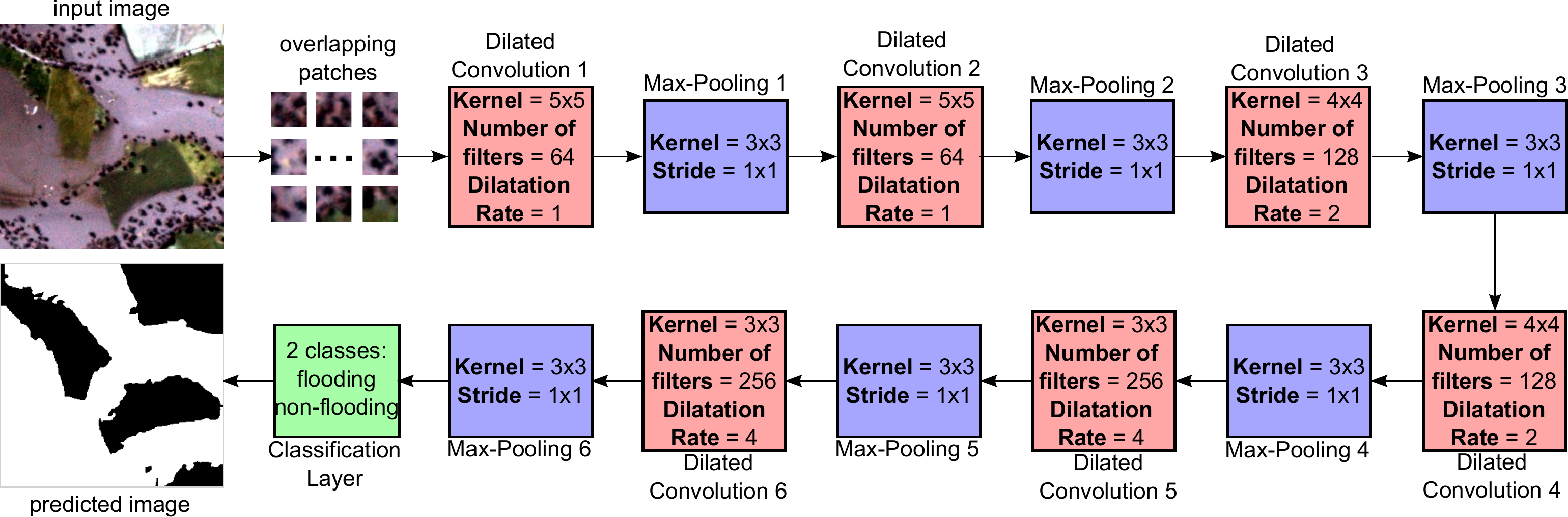}
			\label{dilated2}
		}
		\hspace{1mm}
		\subfloat[Deconvolution ConvNet \#1]{
			\includegraphics[width=\columnwidth]{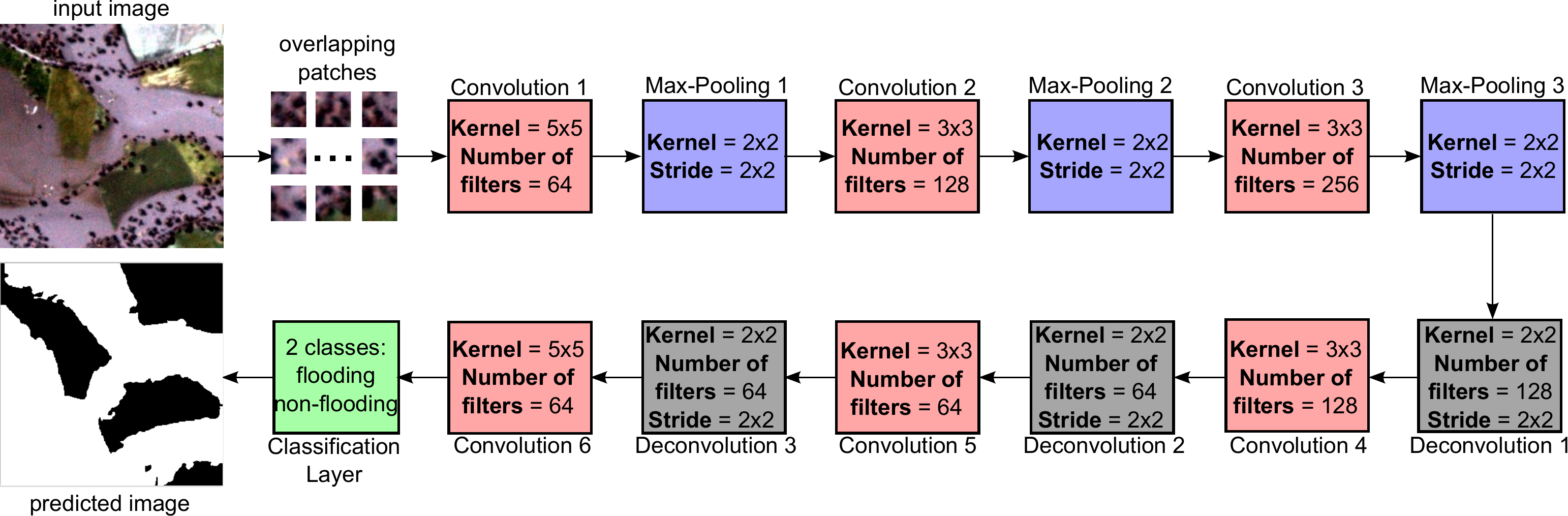}
			\label{deconv1}
		}
		\hspace{1mm}
		\subfloat[Deconvolution ConvNet \#2]{
			\includegraphics[width=\columnwidth]{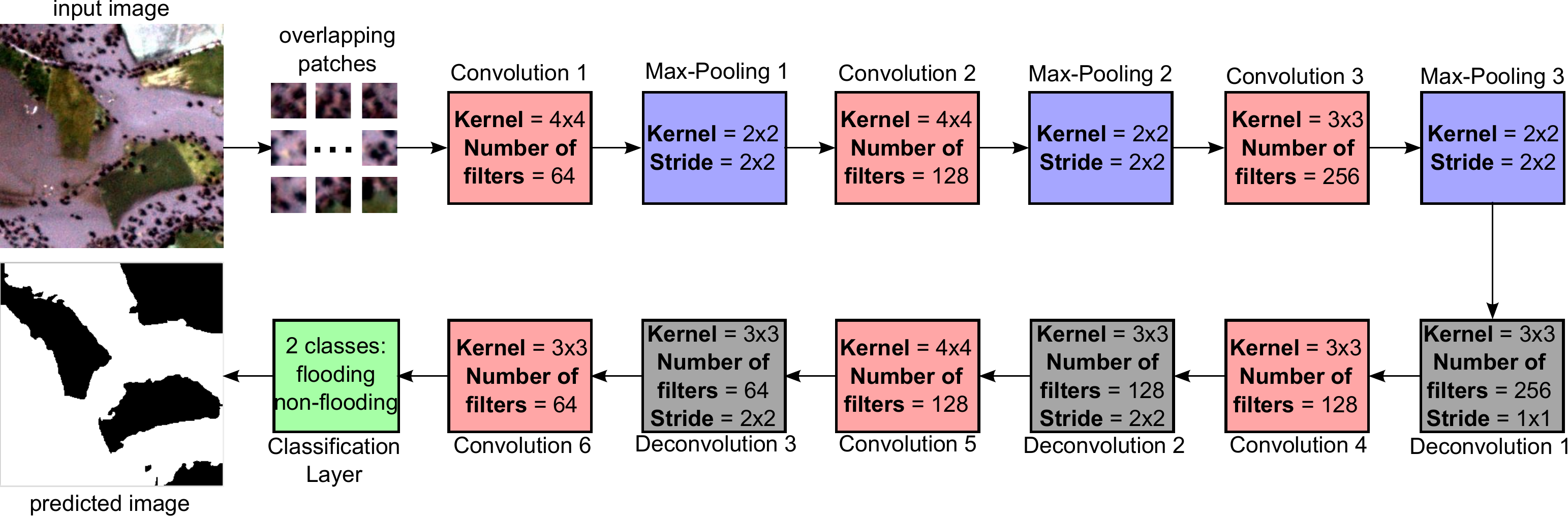}
			\label{deconv2}
		}
		\caption{ConvNet architectures proposed in this work.}
		\label{fig:cnn_archs}
	\end{figure}

	All ConvNets conceived specifically for flooding identification are illustrated in Figure~\ref{fig:cnn_archs}.
	Some use dilated convolutions while others are based on deconvolutional networks.
	Independently of the architecture, Rectified Linear Unit (ReLU)~\cite{nair2010rectified} was used as processing unit for all neurons.
	
	Specifically, two architectures, presented in Figures~\ref{dilated1} and~\ref{dilated2}, are based on the concept of dilated convolutions~\cite{YuKoltun2016}.
	In these layers, the convolution filter is expanded by dilation rate.
	Given this rate, the weights are placed far away at given intervals and the kernel size increases by allowing gaps (or ``holes'') inside their filters.
	Therefore, networks composed of these layers allow the receptive field to expand but preserving the resolution, i.e., \textbf{without downsampling the input data}.
	This procedure represents a great advantage in terms of processing, as well as in terms of learning, since internal feature maps do not lose resolution (and maybe useful information).

	The two remaining networks, presented in Figures~\ref{deconv1} and~\ref{deconv2}, are based on deconvolutional networks~\cite{badrinarayanan2015segnet}.
	This type of network has two modules: the first receives input images, learns the visual features by using standard convolution and pooling layers, and outputs a coarse feature map; while the second receives this map as input, learns to upsample these features by using several deconvolution layers, and outputs a dense prediction map with the same resolution of the original image.
	Both modules work together without distinction and can be trained end-to-end by using standard feedforward and backpropagation algorithms.

	\subsection{Combination} \label{sec:comb}

	We also proposed another strategy to solve the flooding detection task, which aims to exploit the diversity of distinct ConvNets.
	The main premise is that the previous presented ConvNets learn and produce distinct outcomes (dense prediction maps).
	Those differences should make ConvNets complementary to each other.
	Therefore, a clever combination of such outcomes should improve the final prediction map if compared with the ConvNets individual results.
	We propose a combination method using Support Vector Machines (SVM).
	
	\begin{figure}[!t]
		\centering
		\includegraphics[width=\columnwidth]{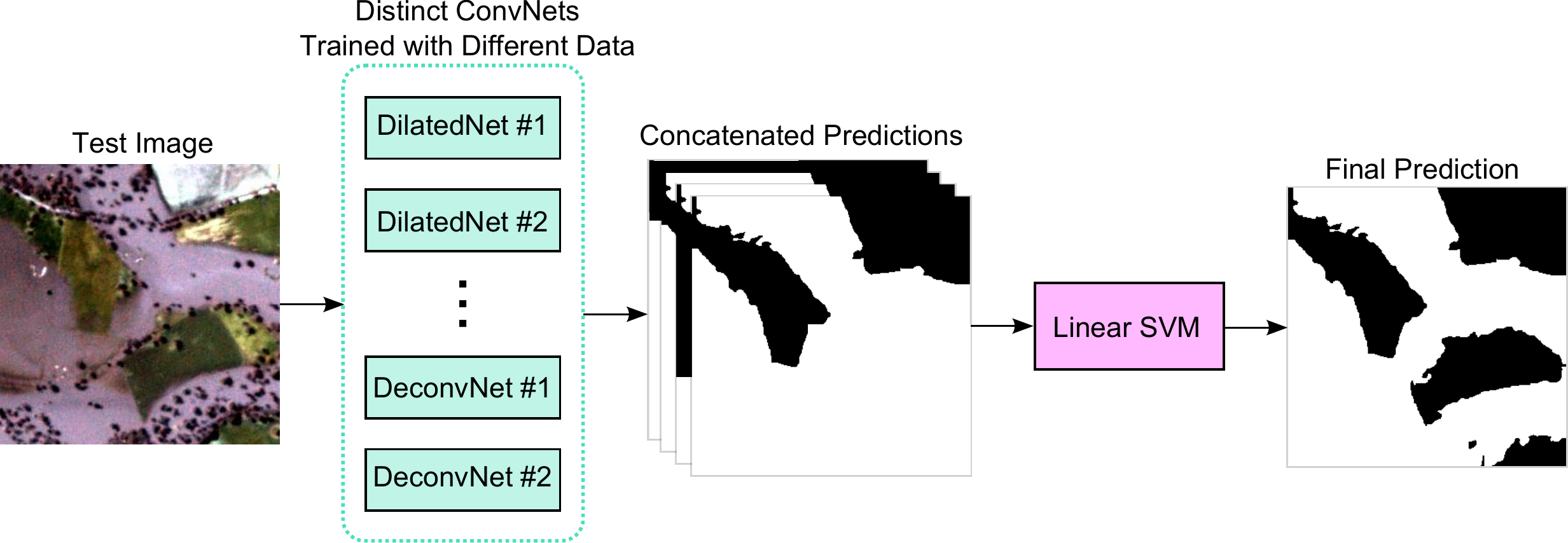}
		\caption{Pipeline for the prediction phase of the combination approach.}
		\label{fig:svm_comb}
	\end{figure}
	
	The proposed method is divided into three main steps:
	(i) {\em extraction:}
	In this phase, an image is processed by all proposed network, which produce distinct outcomes (i.e., different probability or prediction maps).
	All these maps (that have the same resolution of the original input image) are then concatenated creating a feature vector that, in fact, represents the input image.
	(ii) {\em learning:}
	In this step, the SVM receives the aforementioned feature vector, as well as the ground-truth flooding map for all training data.
	Then, it independently processes each pixel of these images, learning which and when each classifier is better; and
	(iii) {\em prediction:}
	This final step receives feature vectors of testing images and, using the trained SVM, outputs the improved prediction map for each test image.
	This final step is illustrated in Figure~\ref{fig:svm_comb}.
	
	\section{Experiments} \label{sec:experiments}

	In this section, we present the experimental setup, including the dataset, evaluation metric, protocol, and baselines.
	
	\subsection{Dataset}
	
	\newcommand{\exSegFigSize}{0.065}
	\begin{figure}[!t]
		\centering
		\subfloat[Location 1]{
			\includegraphics[width=\exSegFigSize\textwidth]{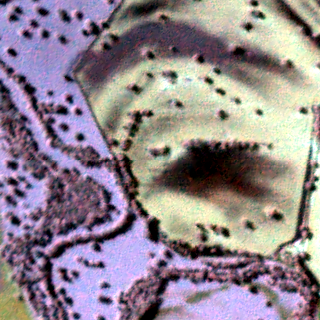}
			\includegraphics[width=\exSegFigSize\textwidth]{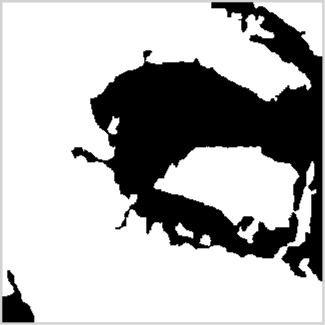}
		}
		\subfloat[Location 2]{
			\includegraphics[width=\exSegFigSize\textwidth]{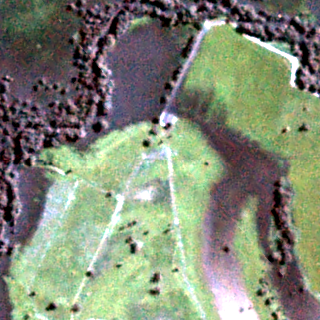}
			\includegraphics[width=\exSegFigSize\textwidth]{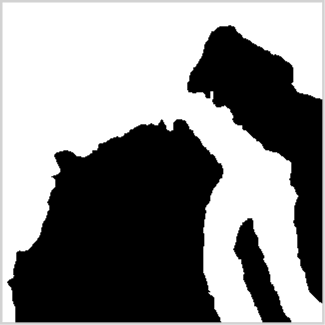}
		}
		\subfloat[Location 3]{
			\includegraphics[width=\exSegFigSize\textwidth]{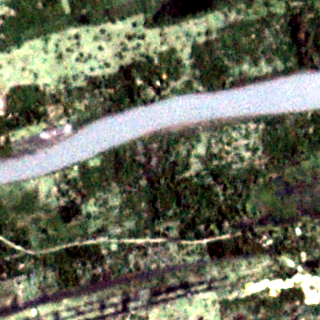}
			\includegraphics[width=\exSegFigSize\textwidth]{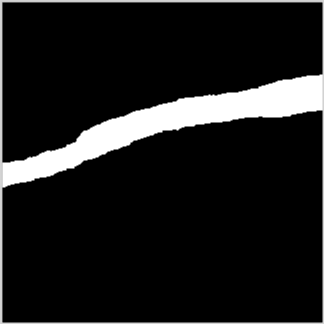}
		}
		\hspace{1mm}
		\subfloat[Location 4]{
			\includegraphics[width=\exSegFigSize\textwidth]{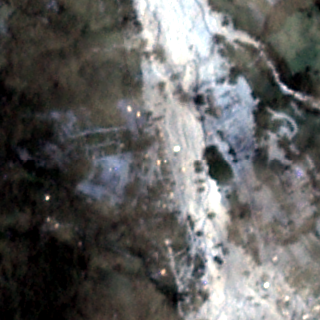}
			\includegraphics[width=\exSegFigSize\textwidth]{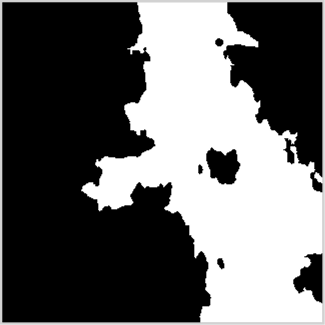}
		}
		\subfloat[Location 5]{
			\includegraphics[width=\exSegFigSize\textwidth]{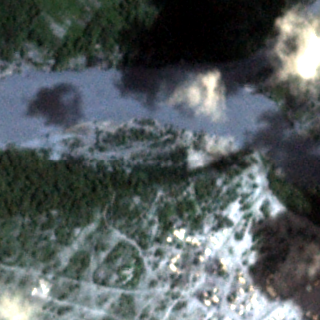}
			\includegraphics[width=\exSegFigSize\textwidth]{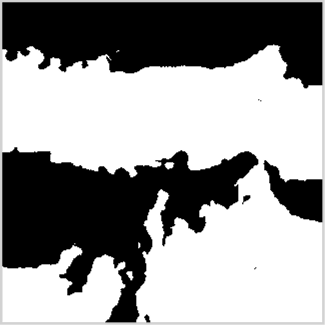}
		}
		\subfloat[Location 6]{
			\includegraphics[width=\exSegFigSize\textwidth]{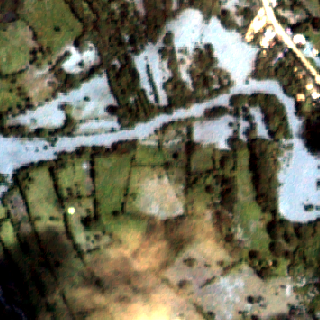}
			\includegraphics[width=\exSegFigSize\textwidth]{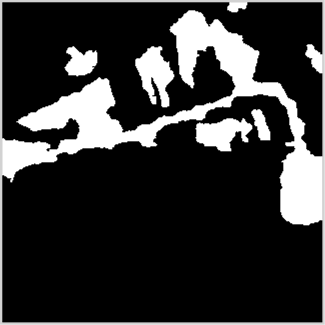}
		}
		\caption{Examples of patches for all locations of the training set with its respective ground-truth, in which white regions refer to flooding areas and black ones correspond to background.}
		\label{fig:training_set_examples}
	\end{figure}
	
	The dataset consists of satellite image patches collected from eight different flooding events June 1st, 2016 to May 1st, 2017.
	Each image patch is composed of four bands (red, green, blue, and near infrared bands) and has resolution of $320\times320$ pixels, with a ground-sample distance (GSD) of 3.7 meters and an orthorectified pixel size of 3 meters~\cite{mediaeval2017satelliteOverview}.
	Each pixel is classified into two classes: background and flooded area.
	While the former class includes everything that is not related to water, the latter, mainly supported by the similarity, includes water in general and does not distinguish between flood and normal surface water (such as river).
	
	The training set is composed of 462 image patches unevenly extracted from \textbf{six} locations.
	Among these images, 92 (20\%) were employed as internal validation set to evaluate the proposed algorithms while the remaining 370 images were used to train the proposed methods.
	Some examples of image patches for each of these locations are presented in Figure~\ref{fig:training_set_examples}.
	Two test sets were released in this dataset:
	the \textbf{Same Locations} test set contains 216 unseen patches unevenly extracted from the same region presented in the training set, while the \textbf{New Locations} test set contains 58 unseen patches extracted from a region not present in the training set.
	Figure~\ref{fig:test_set_results} presents some examples of these test sets.
	Until the submission of this current paper, ground-truth information of the test sets was not released by the organization of the competition.

	\subsection{Experimental Evaluation}
	
	In order to assess the performance of generated segmentation masks for flooded areas in the satellite image patches, the intersection-over-union metric (also known as Jaccard Index), was used.
	The metric measures the accuracy for the pixelwise classification and is defined as $IoU = \frac{TP}{(TP + FP + FN)}$,
	where $TP$, $FP$, and $FN$ are the numbers of true positive, false positive, and false negative pixels, respectively, determined over the whole test set.
	
	\subsection{Experimental Protocol}
	
	First, we trained all presented ConvNets using 50\% overlapping patches of size $25\times25$ extracted from all training images.
	In the prediction phase, we extracted overlapping patches with the same resolution from the testing images and averaged the probabilities outputted by the network.
	Among all networks, the best one is reported as \textbf{ConvNet $25\times25$}.
	
	Another proposed method relied on training the aforementioned ConvNets using larger overlapping patches, with $50\times50$ pixels, also extracted from all training images.
	The motivation behind this strategy is based on the entire context that could be extracted from the input patches to improve the learning process.
	The prediction phase is similar to the previous strategy.
	Considering this configuration, the best network
	is referred, in the next sections, as \textbf{ConvNet $50\times50$}.
	
	The \textbf{Location ConvNets} strategy is based on the idea of creating specialized ConvNets for each flooding event.
	Since the dataset has six distinct flooding event locations, we propose to train a specific Dilated ConvNet \#1 (using patches of $25\times25$) for each location.
	The prediction is similar to the other proposed protocols, except for the fact that, in this case, each ConvNet was used in its respective location.
	For the \textbf{New Locations} test set, we combined the outcomes extracted from each ConvNet (trained specifically for each location) using a linear SVM, as proposed in Section~\ref{sec:comb}.
	
	The \textbf{Fusion-SVM} strategy expands above idea, i.e., an SVM is used to create prediction maps for \textbf{both test sets}, and not only for the \textbf{New Locations} one.
	Based on the premise that distinct ConvNets (trained using different input data) produce distinct (and possibly complementary) outcomes, we propose to combine the predictions extracted from above ConvNets using a linear SVM, as presented in Section~\ref{sec:comb}.
	In this way, the SVM should learn when and how these networks complement each other in order to improve the final performance.

	Another strategy relies on exploiting the diversity of distinct ConvNets by combining outcomes of previous methods using a majority voting scheme, referred as \textbf{Fusion-MV}.

	All proposed methods were created using TensorFlow framework and the code has been made publicly available at~\url{https://github.com/keillernogueira/FDSI/}.
	During training, all the aforementioned protocols used the same hyper-parameters, which were defined based on convergence analyses.
	Specifically, learning rate, weight decay, momentum, and number of epochs are 0.01, 0.0005, 0.9, and 20, respectively.
	After every 5 epochs, the learning rate is reduced 10 times to allow a finer adjustment during the training process.
	
	\subsection{Baselines}
	
	The baselines evaluated in this work were, in fact, approaches proposed for the Flood-Detection in Satellite Images subtask of the 2017 Multimedia Satellite Task.
	An overview of the such methods (which includes state-of-the-art methods, such as Generative Adversarial, Deconvolutional and Fully Convolutional Networks) are presented in Table~\ref{tab:results}.

	\section{Results and Discussion}
	
	In this section, we present the ablation study performed to analyze the diversity of the ConvNets.
	The final results reported by the competition organizers are also presented.
	
	\subsection{Ablation Study} \label{sec:ablation}

	An ablation study, presented in Table~\ref{tab:ablation}, was performed using our validation set in order to evaluate the ConvNets and their diversity.
	In this study, features from all networks (trained using patches of size $25\times25$ and $50\times50$) were extracted and used (individually or in combination) as input for a linear SVM, replicating the proposed \textbf{Fusion-SVM} approach.

	Considering the ConvNet individually, all networks (trained with all events) achieved very similar results.
	Among them, Dilated ConvNet \#1 yielded slightly better results independent of the patch size.
	Therefore, these networks were the ones submitted to the competition and reported in the remaining of this work as ConvNet $25\times25$ and ConvNet $50\times50$.

	Considering the combination, it is possible to see that networks based on same paradigm (first two combinations) do not produce higher performances, suggesting that these networks learn very similar patterns and, therefore, are not complementary.
	However, the combination of distinct networks (dilated with deconvolutional ones, for instance) generates some considerable improvement in the final result for the validation set.
	These results may be due to the distinct pattern extraction process performed by these ConvNets.
	On one hand, dilated networks learn all feasible information without downsampling the input image, which allows them to capture more details mainly in small objects~\cite{YuKoltun2016}.
	On the other hand, deconvolutional networks downsample the input and then learn the final prediction based on a coarse feature map, making them able to give more attention to specific patterns of the image presented in the coarse feature map~\cite{badrinarayanan2015segnet}.
	In addition, the combination of distinct networks trained using the same patch size (all lines in the combination part of the aforementioned table, except the last one) generates slightly different results, with trending to networks trained with $50\times50$ patches.
	This may be justified by the fact that these networks learn distinct features given the different input context and, therefore, provide complementary views about the image content.
	Also, according to the last row, one may notice that the combination of all trained networks leads to an improvement in the final result.
	This increase is comparable to other improvements generated by different network combinations.
	Hence, although we believe that some combinations could produce better results, these would be marginal improvements that do not justify an extensive and costly evaluation of combination scenarios.
	Thus, when concerning the combination strategies, our final submission to the competition was using all previous networks.
	
	\begin{table}[]
		\centering
		\caption{Ablation study over the proposed models.}
		\label{tab:ablation}
		\resizebox{\columnwidth}{!}{
			\begin{tabular}{@{}clrrr@{}}
				\toprule
				\multirow{2}{*}{\textbf{Type}}         & \multicolumn{1}{c}{\multirow{2}{*}{\textbf{Network(s)}}}             & \multicolumn{1}{c}{\multirow{2}{*}{\textbf{Events Considered}}} & \multicolumn{2}{c}{\textbf{Validation}}                                                                                                                                                   \\ \cmidrule(l){4-5} 
				& \multicolumn{1}{c}{}                                                 & \multicolumn{1}{c}{}                                            & \multicolumn{1}{c}{\textbf{\begin{tabular}[c]{@{}c@{}}$25\times25$\\ patches\end{tabular}}} & \multicolumn{1}{c}{\textbf{\begin{tabular}[c]{@{}c@{}}$50\times50$\\ patches\end{tabular}}} \\ 
				\midrule
				\multirow{10}{*}{\textbf{Single}}      & \textbf{Dilated ConvNet \#1}                                         & 1,2,3,4,5,6                                                     & 85.60                                                                                       & 84.98                                                                                      \\
				& \textbf{Dilated ConvNet \#1}                                         & 1                                                               & 62.86                                                                                       &                                                                                             \\
				& \textbf{Dilated ConvNet \#1}                                         & 2                                                               & 66.52                                                                                       &                                                                                             \\
				& \textbf{Dilated ConvNet \#1}                                         & 3                                                               & 51.56                                                                                       &                                                                                             \\
				& \textbf{Dilated ConvNet \#1}                                         & 4                                                               & 35.75                                                                                       &                                                                                             \\
				& \textbf{Dilated ConvNet \#1}                                         & 5                                                               & 37.16                                                                                       &                                                                                             \\
				& \textbf{Dilated ConvNet \#1}                                         & 6                                                               & 35.46                                                                                       &                                                                                             \\
				& \textbf{Dilated ConvNet \#2}                                         & 1,2,3,4,5,6                                                     & 84.21                                                                                       & 83.21                                                                                        \\
				& \textbf{Deconvolution ConvNet \#1}                                   & 1,2,3,4,5,6                                                     & 83.96                                                                                       & 83.04                                                                                       \\
				& \textbf{Deconvolution ConvNet \#2}                                   & 1,2,3,4,5,6                                                     & 83.62                                                                                       & 82.03                                                                                       \\
				\midrule \midrule
				\multirow{12}{*}{\textbf{Combination}} & \textbf{Dilated ConvNet \#1 + \#2}                                   & 1,2,3,4,5,6                                                     & 83.96                                                                                       & 84.69                                                                                       \\
				& \textbf{Deconvolution ConvNet \#1 + \#2}                             & 1,2,3,4,5,6                                                     & 84.66                                                                                       & 84.63                                                                                       \\
				& \textbf{Dilated ConvNet \#1 + Deconvolution ConvNet \#1}             & 1,2,3,4,5,6                                                     & 86.87                                                                                       & 87.32                                                                                       \\
				& \textbf{Dilated ConvNet \#1 + Deconvolution ConvNet \#2}             & 1,2,3,4,5,6                                                     & 86.88                                                                                       & 86.54                                                                                       \\
				& \textbf{Dilated ConvNet \#2 + Deconvolution ConvNet \#1}             & 1,2,3,4,5,6                                                     & 87.03                                                                                       & 87.54                                                                                       \\
				& \textbf{Dilated ConvNet \#2 + Deconvolution ConvNet \#2}             & 1,2,3,4,5,6                                                     & 88.93                                                                                       & 89.10                                                                                       \\
				& \textbf{Dilated ConvNet \#1 + \#2 + Deconvolution ConvNet \#1}       & 1,2,3,4,5,6                                                     & 88.99                                                                                       & 88.91                                                                                       \\
				& \textbf{Dilated ConvNet \#1 + \#2 + Deconvolution ConvNet \#2}       & 1,2,3,4,5,6                                                     & 88.01                                                                                       & 89.52                                                                                       \\
				& \textbf{Dilated ConvNet \#1 + Deconvolution ConvNet \#1 + \#2}       & 1,2,3,4,5,6                                                     & 87.57                                                                                       & 88.95                                                                                       \\
				& \textbf{Dilated ConvNet \#2 + Deconvolution ConvNet \#1 + \#2}       & 1,2,3,4,5,6                                                     & 87.01                                                                                       & 88.65                                                                                       \\
				& \textbf{Dilated ConvNet \#1 + \#2 + Deconvolution ConvNet \#1 + \#2} & 1,2,3,4,5,6                                                     & 88.91                                                                                       & 88.62                                                                                       \\
				& \textbf{ALL (submission)}                                      & 1,2,3,4,5,6                                                     & \multicolumn{2}{c}{88.71}                                                                                                                                                                 \\ 
				\bottomrule
			\end{tabular}
		}
	\end{table}
	
	\subsection{Final Results}
	
\begin{figure}[t!]
		\centering
		\subfloat[Location 1]{
			\label{fig:similar_tasks}
			\includegraphics[width=\exSegFigSize\textwidth]{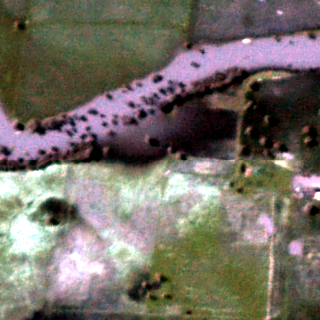}
			\includegraphics[width=\exSegFigSize\textwidth]{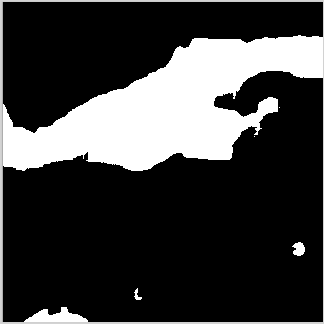}
		}
		\hspace{1mm}
		\subfloat[Location 2]{
			\includegraphics[width=\exSegFigSize\textwidth]{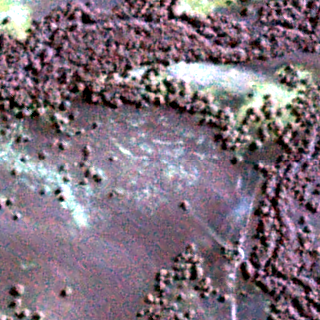}
			\includegraphics[width=\exSegFigSize\textwidth]{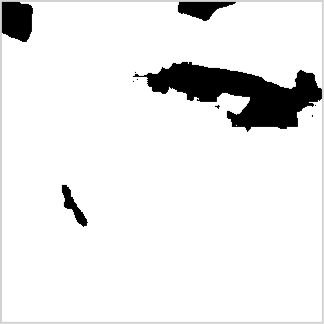}
		}
		\hspace{1mm}
		\subfloat[Location 3]{
			\includegraphics[width=\exSegFigSize\textwidth]{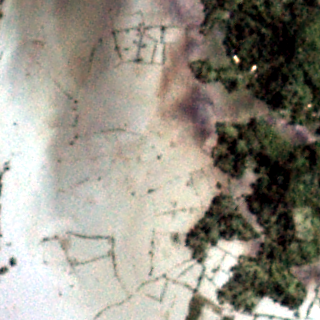}
			\includegraphics[width=\exSegFigSize\textwidth]{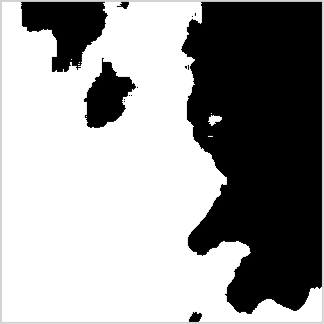}
		}
		\hspace{1mm}
		\subfloat[Location 4]{
			\includegraphics[width=\exSegFigSize\textwidth]{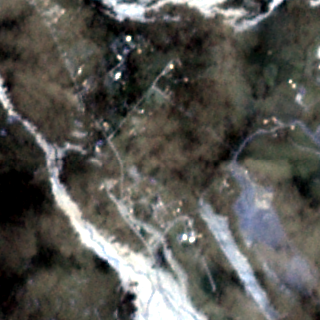}
			\includegraphics[width=\exSegFigSize\textwidth]{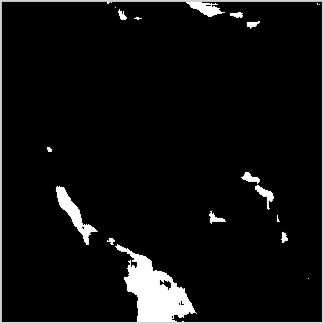}
		}
		\hspace{1mm}
		\subfloat[Location 5]{
			\includegraphics[width=\exSegFigSize\textwidth]{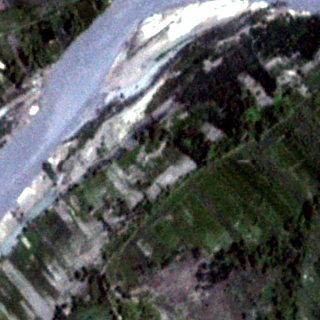}
			\includegraphics[width=\exSegFigSize\textwidth]{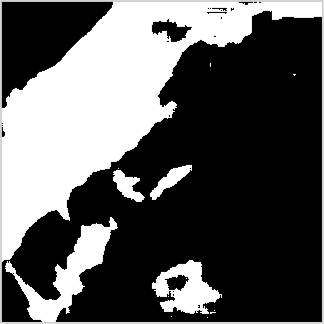}
		}
		\hspace{1mm}
		\subfloat[Location 6]{
			\includegraphics[width=\exSegFigSize\textwidth]{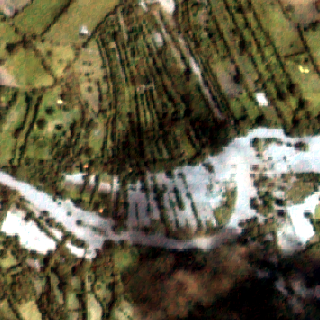}
			\includegraphics[width=\exSegFigSize\textwidth]{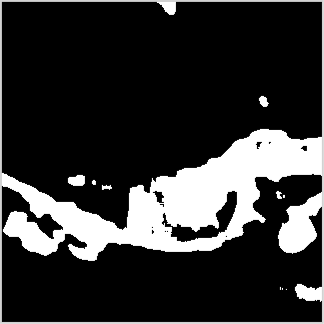}
		}
		\caption{Examples of some test images and the obtained results achieved by using SVM with the aggregated probabilities. White areas refer to flooded regions and black areas correspond to background.}
		\label{fig:test_set_results}
	\end{figure}

	All results for the test sets are presented in Table~\ref{tab:results}.
	These are the official results released by the Mediaeval since no ground-truth for the test set was released yet.
	For both test sets, the best solution was obtained by combining the probabilities of all trained ConvNets using a Linear SVM.
	This technique yielded state-of-the-art results in both test set, outperforming all baselines by, at least, 4\% in the \textbf{Same Locations} test set and 1\% in the \textbf{New Locations} test set (in terms of Jaccard Index).
	Some samples of this results are presented in Figure~\ref{fig:test_set_results}.
	
	For the \textbf{Same Locations} test set, training a network for each location (Location ConvNets) or training a ConvNet with all available data (ConvNet $25\times25$) achieved the same result.
	This may indicate that the proposed architecture can, in fact, extract and interpret all feasible information from the whole data, which is a great advantage given that it reduces the number of networks to train and, consequently, the processing time.
	This conclusion does not hold for the \textbf{New Locations} test set.
	In this set, training a specific network for each location (Location ConvNets) achieved higher performance (aside the Fusion-SVM strategy) when compared to unique networks trained with the whole training set (such as ConvNet $25\times25$ and ConvNet $50\times50$).
	This indicates that specific Location Network can learn details that may not be useful for classification of a known image, but that is important for unseen data, which is the case.
	
	Another relevant outcome is that increasing the size of the input patch (ConvNet $50\times50$) decreases the final result, a conclusion that holds for both datasets.
	We believe that this is because of the amount of training patches generated in each case.
	More specifically, a large amount of data may be used for training with smaller patch sizes, while large patches means less data to train.
	This corroborates with the fact that deep learning really needs a large amount of labeled data to train~\cite{goodfellow2016DLbook}.
	
	Finally, for both sets, the worst result was obtained using the majority voting scheme (Fusion-MV).
	It indicates that Majority Voting is not so robust to aggregate information from multiple networks when they disagree in the classification.
	This fact can be overcome by using a machine learning technique to capture the opinions of the ConvNets.

	Additionally, Figure~\ref{fig:similar_tasks} which (clearly) contains a flooding river, suggests that the proposed method may be also suitable for water segmentation.
	Our hypothesis, left for future validation, is that the proposed methods could be able to distinguish between such classes if the dataset allows a specific training for flood and normal water surface.

	\begin{table}[!t]
		\centering
		\caption{IoU (\%) results of the proposed method and baselines for both test sets. Higher values of IoU indicates better performance.}
		\label{tab:results}
		\resizebox{\columnwidth}{!}{
			\begin{tabular}{@{}lll|rr@{}}
				\toprule
				\multicolumn{2}{c}{\multirow{2}{*}{\textbf{Methods}}} & \multicolumn{1}{c}{\multirow{2}{*}{\textbf{Overview}}} & \multicolumn{2}{c}{\textbf{Test Set}}                           \\
				\cmidrule(l){4-5} 
				\multicolumn{2}{c}{}    & \multicolumn{1}{c}{}   & \multicolumn{1}{c}{\textbf{Same Locations}} & \multicolumn{1}{c}{\textbf{New Locations}} \\
				\midrule
				\multirow{12}{*}{Baselines} & \multirow{2}{*}{WISC~\cite{wisc_mediaeval}} &  NDVI plus SVM-RBF                             & 80                                          & 83                                         \\
				&  & K-Means to cluster and classify                             & 81                                          & 77                                         \\
				\cmidrule(l){2-5}
				& CERTH-ITI~\cite{certh_mediaeval} & Mahalanobis
				dist. with stratified cov.                           
				& 75                                          & 56                                         \\
				\cmidrule(l){2-5}
				& BMC~\cite{bmc_mediaeval} & ResNet-152 and random forest                                 & 37                                          & 40                                         \\
				\cmidrule(l){2-5}
				& \multirow{5}{*}{UTAOS~\cite{utaos_mediaeval}} & Gen. Adv. Net. with 0.78 threshold                            & 82                                          & 73                                         \\
				&  & Gen. Adv. Net. with 0.94 threshold & 80                                          & 70                                         \\
				&  & Gen. Adv. Net. with 0.50 threshold  & 83                                          & 74                                         \\
				&  & Gen. Adv. Net. with 0.35 threshold & 83                                          & 74                                         \\
				&  & Gen. Adv. Net. with 0.12 threshold & 81                                          & 73                                         \\
				\cmidrule(l){2-5}
				& \multirow{3}{*}{DFKI~\cite{dfki_mediaeval}} & VGG13-FCN with RGB data                             & 73                                          & 69                                         \\
				& & VGG13-FCN with RGB and NIR data                              & 84                                          & 70                                         \\
				& & VGG13 adapted to be a DeconvNet                              & 84                                          & 74                                         \\
				\midrule \midrule
				\multirow{5}{*}{Proposed}  & Dilated $25\times25$ & Dilated ConvNet \#1 ($25\times25$ patches)                & 87                                          & 82                                         \\
				& Dilated $50\times50$ & Dilated ConvNet \#1 ($50\times50$ patches)                & 86                                          & 80                                         \\
				& Location ConvNets & Dilated ConvNet \#1 trained per location                   & 87                                          & \textbf{84}                                \\
				& Fusion-SVM & SVM over concatenated predictions                         & \textbf{88}                         		 & \textbf{84}                                \\
				& Fusion-MV  & MV over concatenated predictions                           & 78                                          & 49                                         \\ 
				\bottomrule
			\end{tabular}
		}
	\end{table}

	\section{Conclusion}

	In this paper, we proposed and analyzed the diversity of four distinct deep networks, based on dilated convolutions~\cite{YuKoltun2016} and deconvolution layers~\cite{badrinarayanan2015segnet}, to perform detection of flooding areas in remote sensing images.
	Specifically, each network was analyzed individually and in combination (with an ablation study), allowing a better understanding of their diversity, which resulted in the proposal of a combination method that aims to exploit complementary views of different networks.

	Experimental results showed that the methods are effective and robust.
	We achieved state-of-the-art performance, in terms of Jaccard Index, in a specific dataset proposed for the Flood-Detection in Satellite Images subtask of the 2017 Multimedia Satellite Task.
	The proposed methods outperformed all baselines, winning that subtask challenge.
	Such results show that our proposed approaches are effective and robust to identify flooding areas (independent if it is for a recurrent or atypical event).
	This identification process performed by our proposed algorithms may help authorities keep focus on most vulnerable regions while monitoring forecast inundations, which may aid in coordinate rescues, and help victims.
	As future work, we intend to analyze the proposed method using more datasets to evaluate if it will be able to distinguish between water surfaces.

	\section*{Acknowledgments}
	The authors thank FAPESP (grants \#2013/50169-1, \#2013/50155-0, \#2014/50715-9, \#2014/12236-1, \#2015/24494-8 and \#2016/18429-1), FAPEMIG (APQ-00449-17), CNPq (grant \#312167/2015-6), and CAPES (grant \#88881.145912/2017-01).

	\bibliographystyle{IEEEtran}
	\bibliography{bibliography}
	
\end{document}